\documentclass[10pt,twocolumn,letterpaper]{article}

\usepackage{cvpr}
\usepackage{times}
\usepackage{epsfig}
\usepackage{graphicx}
\usepackage{amsmath}
\usepackage{amssymb}

\usepackage[colorlinks]{hyperref} 
\usepackage{subfiles}
\usepackage{mathtools}

\DeclareMathOperator*{\argmin}{arg\,min}
\DeclareMathOperator*{\argmax}{arg\,max}

\newcommand{\norm}[1]{\left\lVert#1\right\rVert}

\newcommand{\A}{\mathbf{A}}
\newcommand{\f}{\mathbf{f}}
\newcommand{\I}{\mathbf{I}}
\newcommand{\Dec}{\mathcal{D}}
\newcommand{\Enc}{\mathcal{E}}

\newcommand{\Paragraph}[1]{\vspace{-0mm} \noindent \textbf{#1.} \hspace{0mm}}

\newcommand{\SubSection}[1]{\vspace{-1mm} \subsection{#1} \vspace{-1mm}}
\newcommand{\SubSubSection}[1]{\vspace{-3mm} \subsubsection{#1} \vspace{-2mm}}

\usepackage{comment}
\usepackage{graphicx}
\usepackage{multirow}
\usepackage{array}
\usepackage{cite}
\usepackage{amsmath,amssymb,tabularx}
\usepackage[ruled,vlined,linesnumbered]{algorithm2e}
\usepackage{float}
\usepackage{arydshln}
\usepackage{subfigure}

\usepackage{breqn}
\usepackage{xcolor,pifont}

\cvprfinalcopy 

\def\cvprPaperID{5382} 

\newcommand{\firstkey}[1]{\textcolor{red}{\textbf{#1}}}
\newcommand{\secondkey}[1]{\textcolor{blue}{\textbf{#1}}}

\begin{document}

\title{Fully Understanding Generic Objects: \\ Modeling, Segmentation, and Reconstruction}

\author{Feng Liu $\quad$ Luan Tran $\quad$ Xiaoming Liu \\
Michigan State University, East Lansing MI 48824\\
{\tt\small \{liufeng6, tranluan, liuxm\}@msu.edu}
}

\maketitle


\begin{abstract}

Inferring $3$D structure of a generic object from a $2$D image is a long-standing objective of computer vision.
Conventional approaches either learn completely from CAD-generated synthetic data, which have difficulty in inference from real images, or generate $2.5$D depth image via intrinsic decomposition, which is limited compared to the full $3$D reconstruction.
One fundamental challenge lies in how to leverage numerous real $2$D images without any $3$D ground truth. 
To address this issue, we take an alternative approach with semi-supervised learning.
That is, for a $2$D image of a generic object, we decompose it into latent representations of category, shape, albedo, lighting and camera projection matrix, decode the representations to segmented $3$D shape and albedo respectively, and fuse these components to render an image well approximating the input image. 
Using a category-adaptive $3$D joint occupancy field (JOF), we show that the complete shape and albedo modeling enables us to leverage real $2$D images in both modeling and model fitting.
The effectiveness of our approach is demonstrated through superior $3$D reconstruction from a single image, being either synthetic or real, and shape segmentation. 
Code is available at \url{http://cvlab.cse.msu.edu/project-fully3dobject.html}.
\end{abstract}


\section{Introduction}\label{sec:introduction}

\begin{figure}[t]
\centering
\resizebox{\linewidth}{!}{
\includegraphics[trim=5 0 14 0,clip, width=1\linewidth]{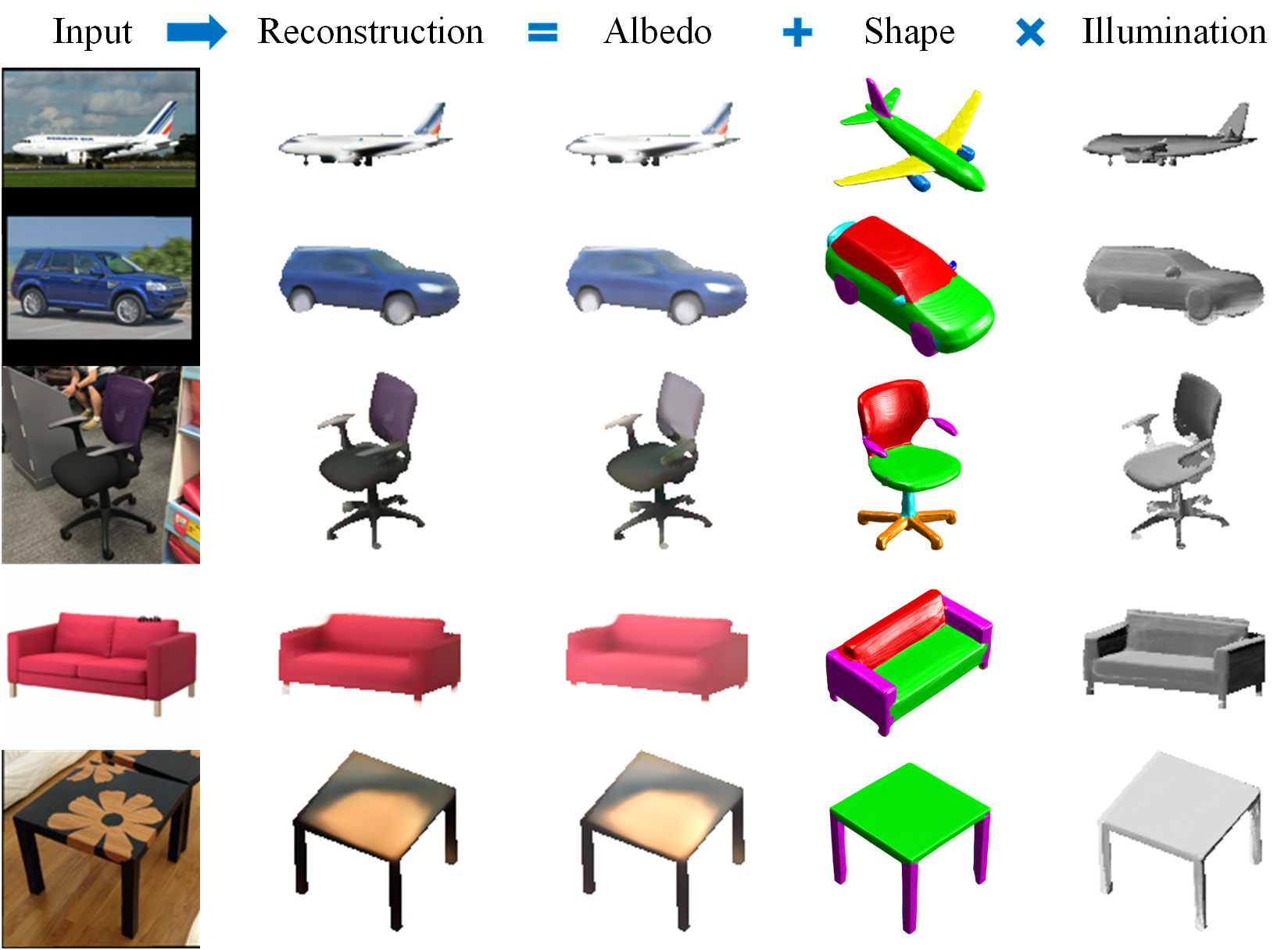}
}
\vspace{-2mm}
\caption{\small Our semi-supervised method learns a universal model of {\it multiple} generic objects. 
During inference, the jointly learnt fitting module decomposes a real $2$D image into albedo, segmented {\it full} $3$D shape, illumination, and camera projection.}
\label{fig:overview}
\vspace{-4.5mm}
\end{figure}

Understanding $3$D structure of objects observed from a single view is a fundamental computer vision problem with applications in robotics, $3$D perception~\cite{brazil2019m3d}, and AR/VR. 
As humans, we are able to effortlessly infer the full $3$D shape when monocularly looking at an object.
%
Endowing machines with this ability remains extremely challenging.
%


With rises of deep learning, many have shown human-level accuracy on $2$D vision tasks, {\it e.g.}, detection~\cite{brazil2019pedestrian,brazil2017illuminating}, recognition~\cite{tran2018representation,tran2017disentangled}, alignment~\cite{zhu2016face}. 
One key reason for this success is the abundance of labeled data. 
%
Thus, the decent performance can be obtained via supervised learning. 
Yet, extending this success to supervised learning for $3$D inference is far behind due to limited availability of $3$D labels.

In this case, researchers focus on using synthetic datasets such as ShapeNet~\cite{chang2015shapenet} 
containing textured CAD models. 
To form image-shape pairs for supervised training, many $2$D images can be rendered from CAD models. 
%
%
However, using synthetic data alone has two drawbacks. Firstly, 
making $3$D object instances is labor intensive and requires  computer graphics expertise, thus not scalable for {\it all} object categories. 
Secondly, the performance of a synthetic-data-trained model often drops on real imagery, due to the obvious domain gap. 
In light of this, \emph{self-supervised} methods can be promising to explore, considering the readily available real-world $2$D images for any object categories, \emph{e.g.,} ImageNet~\cite{russakovsky2015imagenet}. 
If those images can be effectively used in either $3$D object modeling or model fitting, it can have a great impact on $3$D object reconstruction. 



Early attempts~\cite{liu2019soft,tulsiani2017multi} on $3$D modeling from $2$D images in a self-supervised fashion are limited on exploiting $2$D images. Given an image, they mainly learn $3$D models to reconstruct $2$D silhouette~\cite{kato2018neural,liu2019soft}. For better modeling, multiple views of the same object with ground-truth pose~\cite{oechsle2019texture} or keypoint annotations~\cite{kanazawa2018learning} are needed. 
Recent works~\cite{liu2020dist,niemeyer2020differentiable} achieve compelling results by learning from $2$D texture cues via a differentiable rendering.
However, those methods ignore additional monocular cues, \emph{e.g.}, shading, that contain rich $3$D surface normal information.
One common issue in prior works is the lack of separated modeling for albedo and lighting, key elements in real-world image formulation. 
Hence, this would burden the texture modeling for images with diverse illumination variations.  

On the other hand, early work on $3$D modeling for generic objects~\cite{blanz1999morphable,kar2015category,tulsiani2016learning,kanazawa2018learning} often build {\it category-specific} models, where each models intra-class deformation of one category. 
With rapid progress on shape representation, researchers start developing a {\it single universal} model for multiple categories. 
Although such settings expand the scale of training data, it's challenging to simultaneously capture both intra-class and inter-class shape deformations.

We address these challenges by introducing a novel paradigm to jointly learn a completed $3$D model, consisting of $3$D shape and albedo, as well as a model fitting module to estimate the category, shape, albedo, lighting and camera projection parameters from $2$D images of multiple categories (see Fig.~\ref{fig:overview}). 
Modeling albedo, along with estimating the environment lighting condition, enables us to compare the rendered image to the input image in a self-supervised manner. 
Thus, unlabeled real-world images can be effectively used in either $3$D object modeling or learning to fit the model. As a result, it could substantially impact the $3$D object reconstruction from real data. 
Moreover, our shape and albedo learning is conditioned on the category, which relaxes the burden of $3$D modeling for multiple categories. 
This design also enhances the representation power for {\it seen} categories and generalizability for {\it unseen} categories.


A key component in such a learning-based process is a representation  effectively representing both $3$D shape and albedo for diverse object categories. 
Specifically, we propose a category-adaptive $3$D \emph{joint occupancy field} (JOF) conditioned on a category code, to represent $3$D shape and albedo for multiple categories. Using occupancy field as the shape representation, we can express a large variety of $3$D geometry without being tied to a specific topology. 
Extending to albedo, the color field gives the RGB value of the $3$D point's albedo. Modeling albedo instead of texture opens possibility for analysis-by-synthesis approaches, and exploits shading for $3$D reconstruction. Moreover, due to the lack of consistency in meshes' topology, the dense correspondence between $3$D shapes is missing. 
We propose to jointly model the object part segmentation which exploits its implicit correlation with shape and albedo, creating explicit constraints for our model fitting learning.   


In summary, the contributions of this work include:

 $\diamond$ Building a {\it single} model for multiple generic objects. The model fully models segmented $3$D shape and albedo by a \emph{$3$D joint occupancy field}.

 $\diamond$ Modeling intrinsic components enables us to not only better exploit visual cues, but also, leverage real images for model training in a self-supervised manner.
 
 $\diamond$ Introducing a category code into JOF learning, that can enhance the model's representation ability. 

 $\diamond$ Incorporating unsupervised segmentation enables better constraints to fine-tune the shape and pose estimation.

 $\diamond$ Demonstrating superior performance on $3$D reconstruction of generic objects from a single $2$D image.

\section{Prior Work}\label{sec:prior}

\begin{table}[t!]
\newcommand{\tabincell}[2]{\begin{tabular}{@{}#1@{}}#2\end{tabular}}
\newcommand{\greencheck}{\textcolor{green}{\ding{51}}} 
\newcommand{\redX}{\textcolor{red}{\ding{55}}}
\centering
\caption{{\small Comparison of  $3$D object modeling and reconstruction methods. [Keys: CS = category-specific models, SU = a single universal model, Cam = camera parameters, Real data= whether can fine-tune on real-world images self-supervisedly]}}
\vspace{0mm}
\resizebox{1\linewidth}{!}{
\begin{tabular}{l |c| c | c | c | c | c  }
\hline
\multirow{2}{*}{Method} & \multirow{2}{*}{\tabincell{c}{Model \\ type}} & \multirow{2}{*}{\tabincell{c}{\tabincell{c}{Required \\ Cam}}}  & \multicolumn{3}{c|}{Outputs beyond $3$D shapes} & \multirow{2}{*}{\tabincell{c}{Real \\ data}}\\
\cline{4-6}
& & & \tabincell{c}{Texture/Albedo} & \tabincell{c}{Lighting}  & Cam  \\ 
\hline\hline

$3$D-R$2$N$2$~\cite{choy20163d}   & SU  & \redX & \redX & \redX & \redX& \redX\\
PSG~\cite{fan2017point}           & SU  & \redX & \redX & \redX & \redX& \redX\\
AtlasNet~\cite{groueix2018atlasnet} & SU  & \redX & \redX & \redX & \redX& \redX\\
Pixel2Mesh~\cite{wang2018pixel2mesh}& SU  & \redX & \redX & \redX & \redX& \redX\\   
\hdashline
DeepSDF~\cite{park2019deepsdf}      & CS  & \redX & \redX & \redX & \redX& \redX\\
ONet~\cite{mescheder2018occupancy}  & SU  & \redX & \redX & \redX & \redX& \redX\\
IM-SVR~\cite{chen2018learning}     & CS  & \redX & \redX & \redX & \redX& \redX\\ 
\hdashline
\tabincell{c}{Texture Field~\cite{oechsle2019texture}} & CS & \greencheck & Texture & \redX & \redX  & \redX \\ 
PIFu~\cite{saito2019pifu} & CS & \greencheck & Texture & \redX & \redX & \redX\\ 
SRN~\cite{sitzmann2019scene} & CS & \greencheck & Texture & \redX & \redX & \redX\\ 
NeRF~\cite{mildenhall2020nerf} & CS & \greencheck & Texture & \redX & \redX & \greencheck\\
\hdashline
MarrNet~\cite{wu2017marrnet}   & SU  & \redX & \redX & \redX & \redX& \greencheck\\
ShapeHD~\cite{wu2018learning}  & SU  & \redX & \redX & \redX & \redX& \redX\\
F2B~\cite{yao2020front2back}   & SU  & \redX & \redX & \redX & \redX& \redX\\
\hdashline
DRC~\cite{tulsiani2017multi}   & CS  & \greencheck & Texture & \redX & \redX& \greencheck   \\
DIST~\cite{liu2020dist}   & SU  & \greencheck & Texture & \redX & \redX& \greencheck   \\
 Niemeyer \emph{et al.}~\cite{niemeyer2020differentiable}   & SU  & \greencheck & Texture & \redX & \redX& \greencheck   \\
 \hdashline
 CSM~\cite{kulkarni2019canonical,kulkarni2020articulation}   & CS  & \redX & \redX & \redX & \greencheck&
 \greencheck   \\
  CMR~\cite{kanazawa2018learning,goel2020shape}   & CS  & \redX & Texture & \redX & \greencheck&  \greencheck   \\
  UMR~\cite{li2020self}   & CS  & \redX & Texture & \redX & \greencheck&
 \greencheck   \\
\hline

Proposed & SU & \redX & Albedo & \greencheck & \greencheck & \greencheck\\
\hline
\end{tabular}
}
\label{tab:3D_modeling_review}
\vspace{-4mm}
\end{table}

\begin{figure*}[t]
\centering
\includegraphics[trim=0 0 0 0,clip, width=0.99\linewidth]{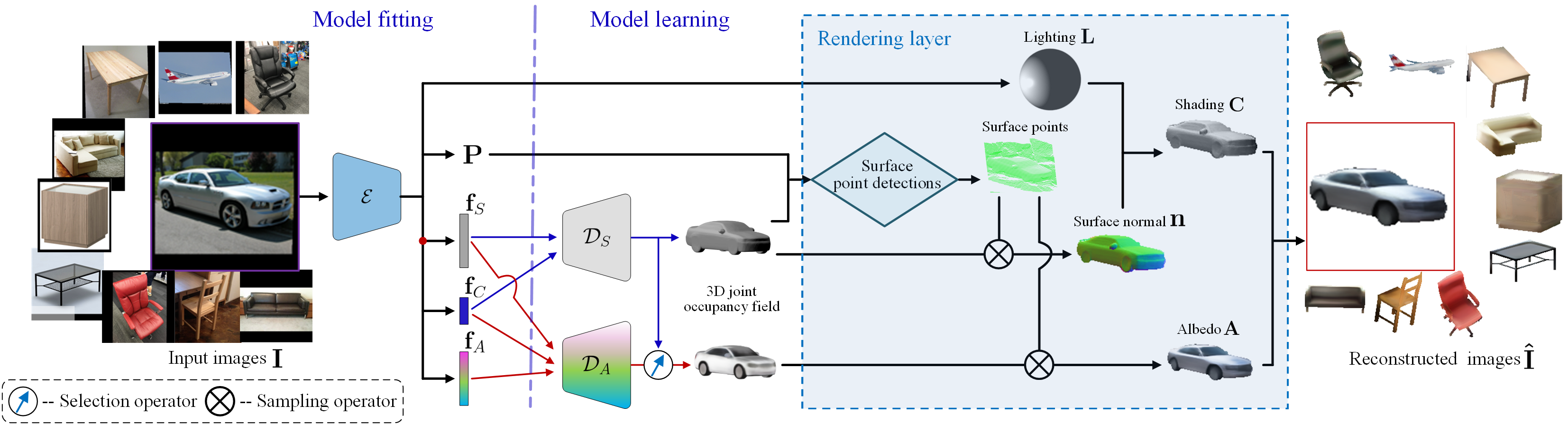}
\vspace{0mm}
\caption{ \small  Semi-supervised analysis-by-synthesis framework jointly learns one image encoder ($\Enc$) and two decoders ($\Dec_S$, $\Dec_A$), with a differentiable rendering layer. Training uses both synthetic and real images, with supervision from class labels and $3$D CAD models, ground truth of synthetic data, and silhouette mask of real data, but not $3$D ground truth of real data.}
\label{fig:architecture}
\vspace{-3mm}
\end{figure*}

\Paragraph{$\mathbf{3}$D Object Representation and Modeling}
Prior works on $3$D object modeling focus more on modeling geometry, based on either point~\cite{qi2017pointnet,qi2017pointnet++,liu2019}, mesh~\cite{groueix2018atlasnet,wen2019pixel2mesh++,meshrcnn}, voxel~\cite{choy20163d,wu2017marrnet,zhang2018learning}, or implicit field~\cite{chen2018learning,mescheder2018occupancy,park2019deepsdf,chibane2020implicit,genova2020local,tretschk2020patchnets,peng2020convolutional}, while less on texture representation.
Current mesh-based texture modeling assumes a predefined template mesh with known topology, limiting to specific object categories, \emph{e.g.}, faces~\cite{tran2019towards,tran2018nonlinear,tran2019on,feng2018prn} or birds~\cite{kanazawa2018learning}.
Recently, several works~\cite{oechsle2019texture,saito2019pifu,sitzmann2019scene,mildenhall2020nerf} adopt the implicit function to regress RGB values in $3$D space, which predicts a complete surface {\it texture}. 
%
By representing a scene as an opaque and textured surface, SRN~\cite{sitzmann2019scene} learns continuous shape and texture representations from posed multi-view images by a differentiable render.  
Mildenhall~\emph{et al.}~\cite{mildenhall2020nerf} represent scenes as neural radiance field allowing novel-view synthesis of more complex scenes. 
However, as summarized in Tab.~\ref{tab:3D_modeling_review}, all these methods assume {\it known} camera parameters or object position, limiting their real-world applicability. 
Further, they are limited to single categories or scenes. 
Our universal model delivers {\it intrinsic $3$D decomposition} for multiple object categories, which map an image to full $3$D shape, albedo, lighting and projection, closing the gap between intrinsic 
image decomposition and practical applications  (Fig.~\ref{fig:overview}).

\Paragraph{Single-view $\mathbf{3}$D Reconstruction}
Learning-based $3$D object modeling~\cite{choy20163d,fan2017point,groueix2018atlasnet,mescheder2018occupancy,chen2018learning,saito2019pifu} can be naturally applied to monocular $3$D reconstruction due to its efficient representation. 
They encodes the input image as a latent vector, from which the decoder reconstructs the pose-neutral $3$D shape. 
However, being trained only on \emph{synthetic} data, many of them suffer from the domain gap.
Another direction is to adopt a two-step pipeline~\cite{wu2017marrnet,wu2018learning,yao2020front2back}, to first recover $2.5$D sketches, 
and then infer a full $3$D shape. 
However, despite $2.5$D eases domain transfer,
they cannot directly exploit $3$D cues from images to mitigate  uncertainty of $3$D representation.
A related line of works~\cite{liu2019learning,liu2020dist,niemeyer2020differentiable,jiang2020sdfdiff} learn to infer $3$D shapes without $3$D label by a differentiable render. 
Another branch of works~\cite{kanazawa2018learning,goel2020shape,li2020self,kulkarni2019canonical,kulkarni2020articulation} learn category-specific, deformable models, or canonical surface mappings based on a template from real images.
However, one common issue among these works is the lack of albedo and lighting modeling, key elements in image formulation, which limits their ability to fully exploit the $2$D image cues.
%

\Paragraph{$\mathbf{3}$D Shape Co-segmentation} 
Co-segmentation operates on a shape collection from a specific category.
Prior works~\cite{yi2018deep,tulsiani2017learning,shu2016unsupervised} develop clustering strategies for meshes, given a handcrafted similarity metric induced by an embedding or graph~\cite{sidi2011unsupervised,xu2010style,huang2011joint}. 
Recently, BAE-NET~\cite{chen2019bae} treats shape co-segmentation as occupancy representation learning, with a branched autoencoder.
BAE-NET is a joint shape co-segmentation and reconstruction network while cares more on segmentation quality. 
Our work extends the branched autoencoder to albedo learning. 
By leveraging correlation between shape and albedo, joint modeling benefits both segmentation and reconstruction.

\Paragraph{$\mathbf{3}$D Morphable Models ($\mathbf{3}$DMMs)} Our framework, as an analysis-by-synthesis approach with $3$D shape and albedo models, is a type of $3$DMMs~\cite{blanz1999morphable}. 
$3$DMMs are widely used to model a single object with small intra-class variation, {\it e.g.}, faces~\cite{blanz1999morphable}, heads~\cite{ploumpis2019combining} or body~\cite{loper2015smpl}.
$3$DMM has not been applied to multiple generic objects due to their large intra-class, inter-class variations and the lack of {\it dense correspondence} among $3$D shapes~\cite{liu2020}. 
To overcome those limitations, we propose a novel $3$D JOF representation to jointly learn a single universal model for multiple generic objects, consisting of both shape and albedo. 
Together with a model fitting module, it allows semi-unsupervised training intrinsic $3$D decomposition network on unlabeled images.

%

\section{Proposed Method}\label{sec:method}
\SubSection{Problem Formulation}

In this work, a generic object is described by three disentangled latent parameters: category, shape and albedo. 
Through two deep networks, these parameters can be decoded into the $3$D shape and albedo respectively.
To have an end-to-end trainable framework, we estimate these parameters along with the lighting and camera projection, via an encoder network, \emph{i.e.}, the fitting module of our model.
Three networks work jointly for the objective of reconstructing the input image of generic objects, by incorporating a physics-based rendering layer, as in Fig.~\ref{fig:architecture}.

Formally, given a training set of $T$ images $\{\mathbf{I}_i \}_{i=1}^T$ of multiple categories, our objective is to learn i)~an encoder $\Enc$: $\mathbf{I}{\rightarrow}\mathbf{P}, \mathbf{L}, \f_C, \f_S, \f_A$ that outputs the projection  $\mathbf{P}$, lighting parameter $\mathbf{L}$, category code $\f_C\in\mathbb{R}^{l_C}$, shape code $\f_S\in\mathbb{R}^{l_S}$, and albedo code $\f_A\in\mathbb{R}^{l_A}$, 
ii) a shape decoder $\Dec_S$ that decodes parameters to a $3$D geometry $\mathbf{S}$, represented by an occupancy field, and 
iii) an albedo decoder $\Dec_A$ that decodes parameters into a color field $\mathbf{A}$, 
with the goal that the reconstructed image by these components ($\mathbf{P}, \mathbf{L},\mathbf{S}, \mathbf{A}$) can well approximate the input.
This objective can be mathematically presented as:
\vspace{-2.5mm}
\begin{equation}
\argmin_{\Enc,\Dec_S, \Dec_A} \sum_{i=1}^T \norm{\hat{\mathbf{I}}_i -  \mathbf{I}_i}_1,
\vspace{-2mm}
\end{equation}
where $\hat{\mathbf{I}} = \mathcal{R} \left( \mathbf{P}, \mathbf{L}, \Dec_S(\f_C, \f_S), \Dec_A(\f_C, \f_S, \f_A) \right)$ is the reconstructed image, and $\mathcal{R}(\cdot,\cdot, \cdot,\cdot)$ is the rendering function. 

\SubSection{Category-adaptive $3$D Joint Occupancy Fields}  

\begin{figure}[t]
\centering
\includegraphics[trim=0 0 0 0,clip, width=\linewidth]{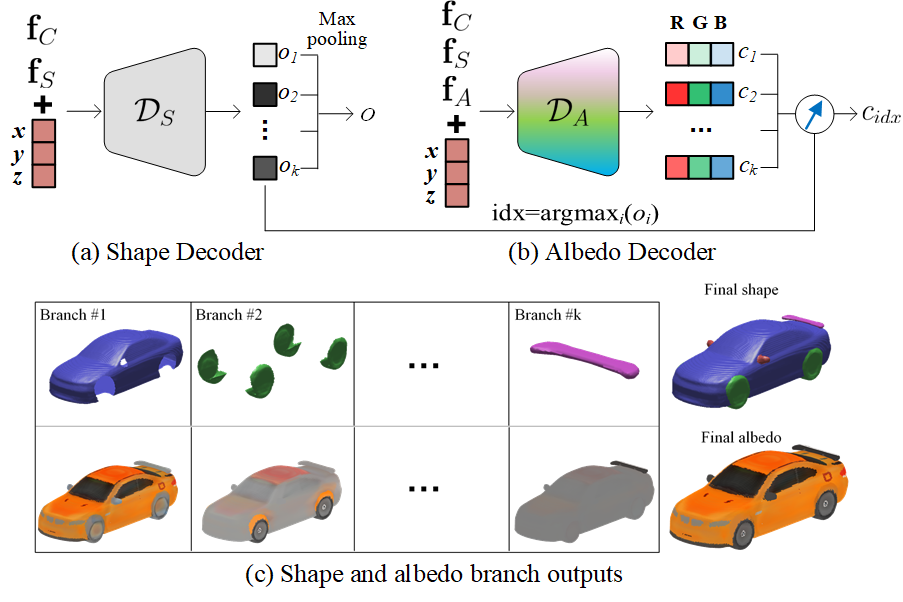}
\vspace{-2mm}
\caption{\small (a) Shape decoder $\Dec_S$ inputs category, shape parameters $\f_C$,  $\f_S$, a spatial point $\mathbf{x} {=} (x,y,z)$, and produces the implicit field for $k$ branches. 
Max pooling of the branch outputs leads to the probability of occupancy $o$.
(b) Albedo decoder $\Dec_A$ receives an additional input $\f_A$ and estimates the albedo of all branches, one of which is selected as the final albedo of $\mathbf{x}$.
(c) Unsupervisedly segmented parts and their albedo match well with intuition.}
\label{fig:decoders}
\vspace{-3mm}
\end{figure}

Unlike $2$D, the community has not yet agreed on a $3$D representation both memory efficient and inferable from data. 
Recently, implicit representations gain popularity as their continuous functions offer high-fidelity surface.
Motivated by this, we propose a $3$D \emph{joint occupancy fields} (JOF) representation to simultaneously model shape and albedo with unsupervised segmentation, offering part-level correspondence for $3$D shapes, as in Fig.~\ref{fig:decoders}.
JOF has three novel designs over prior implicit representations~\cite{chen2018learning,mescheder2018occupancy,park2019deepsdf,oechsle2019texture,chen2019bae}: 1) we extend the idea of unsupervised segmentation~\cite{chen2019bae} from shape to albedo, 2)
we integrate shape segmentation into albedo decoder, guiding segmentation by both geometry and appearance cues, and 3) we condition JOF on the category to model multiple categories.


\Paragraph{Category Code}
Unlike prior implicit representations, we introduce a category code $\mathbf{f}_{C}$ as the additional input to the shape and albedo decoders.
In training, $\mathbf{f}_{C}$ is supervised by a cross-entropy loss using the class label of each image.
In the context of modeling shape deformation of multiple categories, using $\mathbf{f}_{C}$ enables decoders to focus on modeling {\it intra-class} deformations via $\mathbf{f}_{S}$. 
Further, the $\mathbf{f}_{C}$ embedding may generalize to unseen categories too.


\Paragraph{Shape Component}
As adopted from \cite{chen2018learning,mescheder2018occupancy,park2019deepsdf}, each shape is represented by a function, implemented as a decoder network, $\Dec_S$: $ \mathbb{R}^{l_C} {\times}  \mathbb{R}^{l_S} {\times} \mathbb{R}^3   \rightarrow [0,1]$, which inputs a $3$D  location $\mathbf{x}$, category code $\f_C$, shape codes $\f_S$ and outputs its probability of occupancy $o$.
One appealing property is that the surface normal can be computed by the spatial derivative $\frac{\delta \Dec_S}{\delta \mathbf{x}}$ via back-propagation through the network, which is helpful for subsequent tasks such as rendering.

To offer unsupervised part segmentation, we adopt BAE-NET~\cite{chen2019bae} as the architecture of $\Dec_S$. 
It is composed of $3$ fully connected layers
and the final layer is a branched layer that gives the occupancy value for each of $k$ branchs, denoted by $\{o\}_{i=1}^{k}$ in Fig.~\ref{fig:decoders} (a). Finally, a max pooling on the branch outputs the result of the final occupancy. 


\Paragraph{Albedo Component} Albedo component assigns each vertex on the $3$D surface a RGB albedo.
One may use a combination of category and albedo codes to represent a colored shape, \emph{i.e.}, $\Dec_A(\f_C, \f_A, \mathbf{x})$.
However, it puts a redundant burden to $\f_A$ to encode the object geometry, \emph{e.g.}, the position of the tire, and body of a car. 
%
Hence, we also feed the shape code $\f_S$ as an additional input to the albedo decoder, \emph{i.e.}, $\Dec_A$:~$\mathbb{R}^{l_C} {\times} \mathbb{R}^{l_S}  {\times} \mathbb{R}^{l_A} {\times} \mathbb{R}^3  \rightarrow \mathbb{R}^{3}$ (Fig.~\ref{fig:decoders}(b)).




Inspired by the design of $\Dec_S$, we propose to estimate the albedo for $k$ branches $\{c\}_{i=1}^{k}$.
For each $\mathbf{x}$, the final albedo is $c_{idx}$, where $idx{=} \argmax_i (o_i)$ is the index of segment where  $\mathbf{x}$ belongs to (Fig.~\ref{fig:decoders}(c)). 
This novel design integrates shape segmentation into albedo learning, benefiting both segmentation and reconstruction (Tab.~\ref{tab:segmentation}). 
The key motivation is that, different parts of an object often differ in shape {\it and/or} albedo, and thus both shall guide the segmentation. 

\SubSection{Physics-based Rendering}
\label{sec:rendering}

To render an image ($W \times H$ pixels) from shape, albedo, as well as lighting parameters $\mathbf{L}$ and projection $\mathbf{P}$, we first find a set of $W \times H$ $3$D surface points corresponding to the $2$D pixel. Then the RGB color of each pixel is computed via a lighting model using lighting  $\mathbf{L}$ and decoder outputs.

\Paragraph{Camera Model}
We assume a full perspective camera model. Any spatial points $\mathbf{x}$ in the world space can be projected to camera space by a multiplication between a $3 {\times} 4$ full perspective projection matrix $\mathbf{P}$ and its homogeneous coordinate: $\mathbf{u} = \mathbf{P} \left[ \mathbf{x}, 1 \right]^T$, $\mathbf{u}=[u\cdot d,v\cdot d,d]^T$, where $d$ is the depth value of image coordinate $(u,v)$. 
Essentially, $\mathbf{P}$ can be extended to a $4{\times}4$ matrix. 
By an abuse of notation in homogeneous coordinates, relation between $3$D points $\mathbf{x}$ and its camera space projection $\mathbf{u}$ can be written as:
\vspace{-2mm}
\begin{equation}
\mathbf{u} = \mathbf{P} \mathbf{x}, \hspace{3mm} \text{and} \hspace{3mm} \mathbf{x} = \mathbf{P}^{-1} \mathbf{u}.
\vspace{-2mm}
\end{equation}

\Paragraph{Surface Point Detection}\label{sec:point_select}
To render a $2$D image, for each ray from the camera to the pixel $j = (u,v)$, we select one ``surface point''. Here, a surface point is defined as the first interior point ($\Dec_S(\mathbf{x})>\tau$), or the exterior point with the largest $\Dec_S(\mathbf{x})$ in case the ray doesn't hit the object. 
For efficient training, instead of finding exact surface points, we approximate them via Linear or Linear-Binary search.  Intuitively, with the distance margin error of $\epsilon$, in Linear search, along each ray we evaluate $\Dec_S(\mathbf{x})$ for all spatial point candidates $\mathbf{x}$ with a step size of $\epsilon$. In Linear-Binary search, after the first interior point is found, as $\Dec_S(\mathbf{x})$ is a continuous function, a Binary search can be used to better approximate the surface point.
With the same computational budget, Linear-Binary search leads to better approximation of surface points, hence higher rendering quality.
The  search algorithm is  detailed in the supplementary material (\emph{Supp.}).

\Paragraph{Image Formation}
We assume purely Lambertian surface reflectance and distant low-frequency illumination. Thus, the incoming radiance can be approximated via Spherical Harmonics (SH) basis functions $\mathbf{H}_b: \mathbb{R}^3 \rightarrow \mathbb{R}$, and controlled by coefficients $\mathbf{L}=\{\gamma_b\}_{b=1}^{3B^2}$.
At the pixel $j$ with corresponding surface point $\mathbf{x}_j$, the image color  is computed as a product of albedo $\mathbf{A}_j$ and shading $\mathbf{C}_j$:


\vspace{-5mm}
\begin{equation}
\I_j {\,=\,} \A_j \cdot \mathbf{C}_j {\,=}  \Dec_A(\mathbf{x}_j) \cdot \sum_{b=1}^{B^2} \gamma_b \mathbf{H}_b \left( \sigma \left( \frac{\delta \Dec_S(\mathbf{x}_j ) }{\delta \mathbf{x}_j} \right) \right),
\label{eqn:render}
\vspace{-2mm}
\end{equation}
where $\mathbf{n}_j{=}\sigma \left( \frac{\delta \Dec_S(\mathbf{x}_j ) }{\delta \mathbf{x}_j} \right)$ is the surface normal direction at $\mathbf{x}_j$, $L_2$-normalized by function $\sigma()$. We use $B{=}3$ SH bands, which leads to $B^2{=}9$ coefficients for each color channel. 

\SubSection{Semi-Supervised Model Learning}
While our model is designed to learn from real-world images, we benefit from pre-training shape and albedo with CAD models, given inherent ambiguity in inverse tasks. 
We first describe learning from images self-supervisedly, and then pre-training from CAD models with supervision.


\SubSubSection{Self-supervised Joint Modeling and Fitting}
Given a set of $2$D images without ground truth $3$D shapes, we define the loss as ($\lambda_i$ are the weights):
\vspace{-1mm}
\begin{equation}
\argmin_{\Enc, \Dec_A} \mathcal{L}_3 = \mathcal{L}_{\text{img}} + \lambda_1 \mathcal{L}_{\text{sil}} + \lambda_2 \mathcal{L}_{\text{fea-const}}  + \lambda_3 \mathcal{L}_{\text{reg}},
\vspace{-1mm}
\label{eqn:trainloss}
\end{equation}
where $\mathcal{L}_{\text{img}}$ is the photometric loss, $\mathcal{L}_{\text{sil}}$ enforces silhouette consistency, 
$\mathcal{L}_{\text{fea-const}}$ is the local feature consistency loss, and $\mathcal{L}_{\text{reg}}$ includes two regularization terms ($\mathcal{L}_{\text{alb-const}}$, $\mathcal{L}_{\text{bws}})$.

\Paragraph{Silhouette Loss}
Given the object’s silhouette $\mathbf{M}$, obtained by a segmentation method~\cite{qin2020u2}, we define the loss as:

\vspace{-4mm}
\begin{equation}
\resizebox{0.9\hsize}{!}{$
\mathcal{L}_{\text{sil}} = \frac{1}{W\times H} \sum_{j=1}^{W{\times}H} \norm{ \Dec_S(\Enc_C(\mathbf{I}), \Dec_S(\Enc_S(\mathbf{I}), \Enc_P(\mathbf{I})^{-1}\mathbf{u}_j) -  o_j }_1,$}
 \label{eqn:L_sil}
\end{equation}
\noindent where $\Enc_C, \Enc_S, \Enc_P$ are parts of the encoder that estimate $\mathbf{f}_C$, $\mathbf{f}_S$ and $\mathbf{P}$ respectively and the three inputs to $\Dec_S$ are $\f_C$, $\f_S$ and  $\mathbf{x}_j$.
With the occupancy field, the occupancy value $o_j$ is $0.5$ if $\mathbf{M}_j=1$, otherwise $o_j=0$. 
Here, we also analyze how our silhouette loss differs from
prior work. If a 3D shape is represented as a mesh, there is no gradient when comparing two binary masks, unless the predicted silhouette is expensively approximated as in Soft rasterizer~\cite{liu2019soft}.
If the shape is represented by a voxel, the loss can provide gradient to adjust voxel occupancy predictions, but not the object orientation~\cite{tulsiani2017multi}.
Our loss can update both shape occupancy field and camera projection estimation (Eqn.~\ref{eqn:L_sil}).

\Paragraph{Photometric Loss} To enforce similarity between our reconstruction and input, we use a $L_1$ loss on the foreground:
\vspace{-2mm}
\begin{equation}
\mathcal{L}_{\text{img}} = \frac{1}{|\mathbf{M}|}\norm{ (\mathbf{\hat{I}} - \mathbf{I})  \odot \mathbf{M} }_1,
\label{eqn:reconLoss} \vspace{-1mm}
\end{equation}
where $\odot$ is the element-wise product. 
%

\begin{figure}[t]
\centering
\resizebox{\linewidth}{!}{
\includegraphics[trim=13 0 0 0,clip,width=1\linewidth]{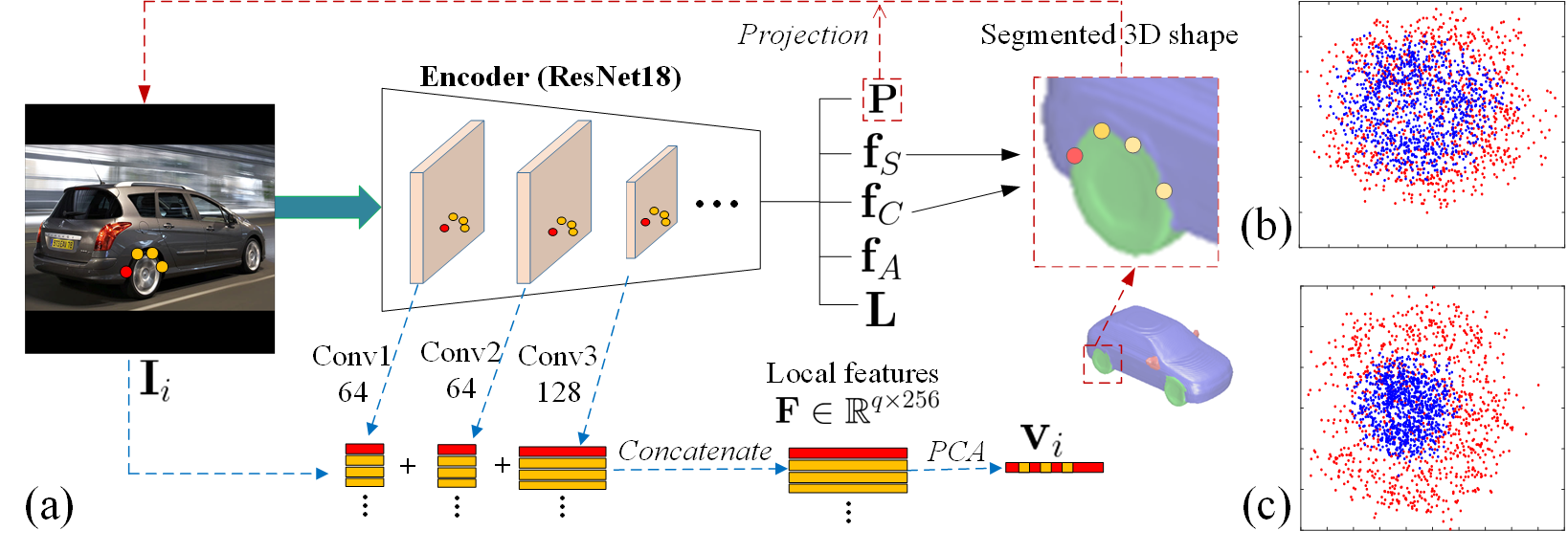}
}
\vspace{-2mm}
\caption{\small{(a) Local feature extraction. For an image $\mathbf{I}_{i}$, part segmentation allows   selecting and projecting $3$D boundary points onto $\mathbf{I}_{i}$. Using their $2$D locations to sample the first $3$ feature maps of the encoder results in the set of local features $\mathbf{F}$, whose eigenvector $\mathbf{v}_{i}$ is used in $\mathcal{L}_{\text{fea-const}}$. 
t-SNE plots of $\mathbf{v}_{i}$ from $1{,}000$ car images using the models trained without (b) or with (c) $\mathcal{L}_{\text{fea-const}}$. Blue and red are $\mathbf{v}_{i}$ of boundary pixels and randomly sampled pixels respectively. While distributions of random pixels remain scattered, $\mathcal{L}_{\text{fea-const}}$ 
helps boundary pixels to have more similar feature distribution, thus better semantic correspondence across images.}}

\label{fig:local_fea}
\vspace{-2mm}
\end{figure}

\Paragraph{Local Feature Consistency Loss}
Our decoders \emph{unsupervisedly} offer part-level correspondence via learnt segmentation (Fig.~\ref{fig:decoders}), with which
we assume that the boundary pixels of adjacent segments in one image have a similar \emph{distribution} of appearance as another image of the same category. 
This assumption leads to a novel loss function (Fig.~\ref{fig:local_fea}).

For one segmented $3$D shape, we first select $q$ boundary points $\mathbf{U}_{3D}\in \mathbb{R}^{q\times 3}$ from all pairs of adjacent segments based on branches of $\Dec_S$, \emph{i.e.}, a point and its spatial neighbor shall trigger different branches. 
These $3$D points are projected to the image plane $\mathbf{U}_{2D}\in \mathbb{R}^{q\times 2}$ via estimated  $\mathbf{P}$. 
Similar to~\cite{xu2019disn}, we retrieve features from feature maps via the location $\mathbf{U}_{2D}$ and form the local features $\mathbf{F}\in \mathbb{R}^{q \times 256}$, where $256$ is the total feature dimension of $3$ layers. 
Finally, we calculate the largest eigenvector $\mathbf{v}$ of the covariance matrix $(\mathbf{F}-\mu)^T(\mathbf{F}-\mu)$ ($\mu$ is the row-wise mean of $\mathbf{F}$), which describes the largest feature variation of $q$ points.
Despite two images of the same category may differ in colors, we assume there is similarity in their respective major variations. 
Thus, we define the local feature consistency loss as:
\vspace{-2mm}
\begin{equation}
\mathcal{L}_{\text{fea-const}} = \frac{1}{|B|}\sum_{(i,j)\in B} \norm{\mathbf{v}_i-\mathbf{v}_j}_1,
\vspace{-2mm}
\end{equation}
where $B$ is a training batch of the same category. 
This loss drives the semantically equivalent boundary pixels across multiple images to be projected from the same $3$D boundary adjoining two $3$D segments, thus improving pose and shape estimation. 

\Paragraph{Regularization Loss} We define two regularizations: 

\noindent\emph{Albedo local constancy}: assuming piecewise-constant albedo~\cite{land1971lightness}, we enforce the gradient sparsity in two directions~\cite{shu2017neural}:
$\mathcal{L}_{\text{alb-const}} =  \sum_{t \in \mathcal{N}_j} \omega (j, t) \norm{ \mathbf{A}_j - \mathbf{A}_t }_2^p$,
where $\mathcal{N}_j$ represents pixel $j$'s  $4$ neighbor pixels.
Assuming that pixels with the same chromaticity (\emph{i.e.}, $\mathbf{c}_j = {\mathbf{I}_j}/{|\mathbf{I}_j|}$)  are more likely to have the same albedo, we set the weight $\omega (j,t) = e^{-\alpha \norm{ \mathbf{c}_j - \mathbf{c}_t }}  $, where the color is referred of the input image.
We set $\alpha=15$ and $p=0.8$ as in~\cite{meka2016live}.

\noindent\emph{Batch-wise White Shading}: Similar to ~\cite{shu2017neural}, to prevent the network from generating arbitrary bright or dark shading, we use a batch-wise white shading constraint:
$\mathcal{L}_{\text{bws}} = \norm{\frac{1}{m} \sum_{j=1}^m \mathbf{C}_j^{(r)} - c}_1$,
where $\mathbf{C}_j^{(r)}$ is a red channel diffuse shading of pixel $j$. $m$ is the total number of foreground pixels in a mini-batch.
$c$ is the average shading target, which is set to $1$.
The same constraint is applied to other channels.

\SubSubSection{Supervised Learning with Synthetic Images} 

Before self-supervision, we pre-train with CAD models and synthetic data, vital for converging to faithful solutions.

\Paragraph{Pre-training Shape and Albedo Decoder}
For auto-encoding $3$D shape and albedo, we adopt a $3$D CNN as encoder $\Enc'$ to extract category, shape and albedo codes $\f_{C}$, $\f_{S}$, $\f_{A}$ from a $64^3{\times}3$ \emph{colored} voxel. As in Fig.~\ref{fig:pretrain},
given a dataset of CAD models, a model (with class label $y$) can be represented as a colored $3$D occupancy voxel $\mathbf{V}$. Equivalently, it can also be represented by $K$ spatial points $\mathbf{x}\in\mathbb{R}^3 $ and their occupancy $o$,  albedo $c$. 
We define the following loss:

\begin{figure}[t]
\centering
\includegraphics[trim=0 0 0 0,clip, width=0.99\linewidth]{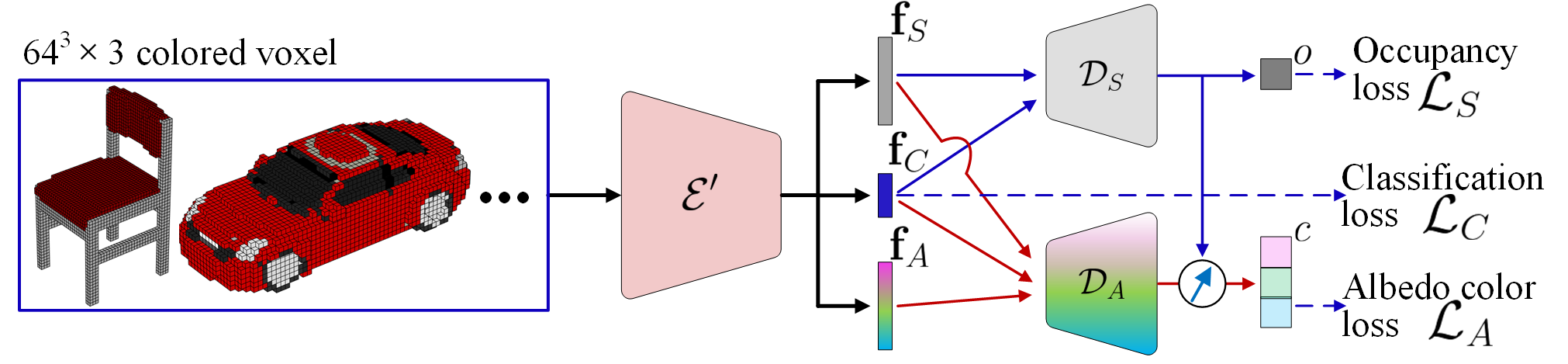}
\caption{ \small Colored $3$D voxel encoder $\mathcal{E'}$ and decoders pre-training. 
}
\label{fig:pretrain}
\vspace{-3mm}
\end{figure}

\vspace{-2mm}
\begin{equation}
    \argmin_{\Dec_S, \Dec_A, \Enc' } \mathcal{L}_1 = \mathcal{L}_S + \mathcal{L}_A + \mathcal{L}_C,
\label{eqn:pretrain1}
\vspace{-2mm}
\end{equation}
where $\mathcal{L}_S{=}\sum_{j=1}^K \norm{ \Dec_S( \Enc'_C(\mathbf{V}), \Enc'_S(\mathbf{V}), \mathbf{x}_j) - o_j}_2^2$, $\mathcal{L}_A{=}\sum_{j=1}^K \norm{ \Dec_A( \Enc'_C(\mathbf{V}), \Enc'_S(\mathbf{V}), \Enc'_A(\mathbf{V}), \mathbf{x}_j)- c_j}_2^2$, and $\mathcal{L}_C$ is cross-entropy loss for class label $y$. 
Note that training $\Enc'$ is necessary to learn valid distributions of $\f_C$, $\f_S$, $\f_A$, although $\Enc'$ is discarded after this pre-training step.


\Paragraph{Pre-training Image Encoder}
Given a CAD model, we render multiple images of the same object with different poses and lighting, each forming a triplet of voxel, image and ground truth projection $( \mathbf{V}, \I,  \mathbf{\widetilde{P}})$.
These synthetic data can supervise the pre-training of encoder  $\Enc$ by minimizing the $\mathcal{L}_2$ below, where the ground truth shape and albedo parameters are obtained by feeding voxel $\mathbf{V}$ into $\Enc'$,
\vspace{-3mm}
\begin{equation}
\resizebox{0.99\hsize}{!}{$
\mathcal{L}_2 = \mathcal{L}_{\text{img}}  + \sum_{X \in \{C, S, A\} } \lambda_X \norm{ \Enc_X(\mathbf{I})-\Enc'_X(\mathbf{V}) }^2_2  \nonumber + \lambda_P \norm{ \Enc_P(\mathbf{I})-\mathbf{\widetilde{P}} }^2_2$}.
\label{eqn:pretrain2}
\vspace{-1mm}
\end{equation}

\SubSection{Implementation and Discussion}


Our training process contains three steps: 1) $\Dec_S$, $\Dec_A$ and $\Enc'$ are pre-trained on colored voxels and corresponding sampled point-value pairs  (Eqn.~\ref{eqn:pretrain1}); 
2) $\Enc$ is pre-trained with synthetic images by minimizing $\mathcal{L}_2$; 3) $\Enc$ and $\Dec_A$ are trained using real images (Eqn.~\ref{eqn:trainloss}).
We empirically found that, Step $3$ training has incremental gain when updating the shape decoder. But it significantly improves the generalization ability of our encoder on fitting model to real images. Thus, we opt to freeze the shape decoder after Step $1$.
For more details about the training setting, please refer to \emph{Supp.}. 

One key enabler of our learning with real images is the \emph{differentiable} rendering layer. 
For the rendering function of Eqn.~\ref{eqn:render}, one can compute partial derivatives over $\mathbf{L}$, over $\mathbf{P}$ since $\mathbf{x} = \mathbf{P}^{-1} \mathbf{u}$, over $\f_C$, $\f_S$, $\f_A$ as they are the inputs of  $\Dec_S$, $\Dec_A$, and over the network parameters of $\Dec_S$, $\Dec_A$. 
However, although the derivative over $\mathbf{x}_j$ can be computed, the surface point search process is not differentiable.



%

\section{Experimental Results}\label{sec:exp}

\Paragraph{Data} We use the ShapeNet Core v1~\cite{chang2015shapenet} for  pre-training in Steps $1$-$2$. 
Following the settings of~\cite{choy20163d,wang2018pixel2mesh,mescheder2018occupancy}, we use CAD models of $13$ categories and the same training/testing split. 
While using the same test set, we render training data ourselves, adding lighting and pose variations. 
We use real images of Pascal $3$D${+}$~\cite{xiang2014beyond} in Step $3$ training. 
We select $5$ categories (plane, car, chair, couch and table) which overlap with $13$  categories in synthetic data. 
   
\Paragraph{Metrics} We adopt standard $3$D reconstruction metrics: F-score~\cite{knapitsch2017tanks} and Chamfer-$L_1$ Distance (CD). 
 Following~\cite{wang2018pixel2mesh}, we calculate precision and recall by checking the percentage of points in prediction or ground-truth that can find the nearest neighbor from the other within a threshold $\tau$. 
Following~\cite{mescheder2018occupancy}, we randomly sample $100k$ points from ground-truth and estimated meshes, to compute CD.

\SubSection{Ablation and Analysis}\label{sec:ab_study}

\begin{table}
\centering
\newcommand{\tabincell}[2]{\begin{tabular}{@{}#1@{}}#2\end{tabular}}
\vspace{0mm}
\caption{\small Reconstruction comparison between category-specific (CS) and single universal (SU) models on $13$ ShapeNet categories.}
\vspace{0mm}
\centering
\resizebox{0.99\linewidth}{!}{
\begin{tabular}{l| c| c| c }
\hline
Model &  CS   &  
{SU (w/o category code)} &
{SU (w.~category code)}\\
\hline 
\hline 
Average CD $\downarrow$ &  $0.149$ &  $0.193$ &  $0.168$  \\
\hline 
\end{tabular}
}
\label{tab:model_type}
\vspace{-4mm}
\end{table}

\Paragraph{Single vs. Category-specific Models} We compare two set of models on ShapeNet data: category-specific (CS) models and single universal (SU) models. 
CS models specialize for each particular class, which of course has better reconstruction quality (Tab.~\ref{tab:model_type}), and may define upper bound performance for SU.
Further, we ablate the single universal model with or without category code $\f_C$. Clearly, the one with category code performs better, which shows that the category code does relax the burden on decoders and enable the decoders focus on the intra-class shape deformations.

\begin{figure}[t]
\centering
\includegraphics[trim=10 0 0 0, clip,width=\linewidth]{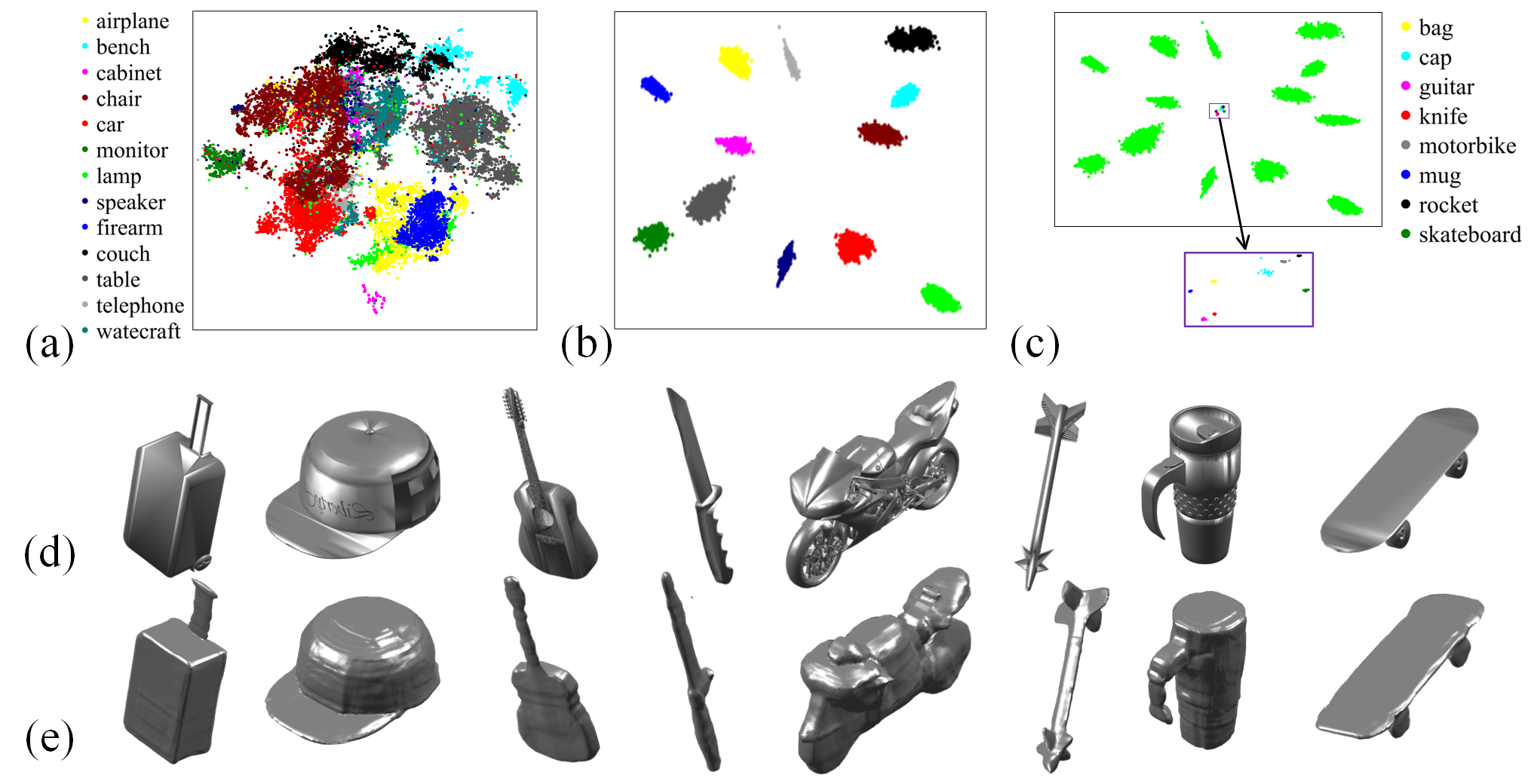}
\vspace{-2mm}
\caption{\small (a), (b) show t-SNE plots of $\mathbf{f}_{S}$ and $\mathbf{f}_{C}$ respectively. (c)~t-SNE plot of the estimated $\mathbf{f}_{C}$ of $8$ unseen classes. (d)~The ground-truth shapes of the testing unseen classes. (e)~The best shapes our shape decoder can reconstruct. \textbf{No encoder is involved.}}
\label{fig:latent_}
\vspace{0mm}
\end{figure}

\Paragraph{$\mathbf{f}_{S}$, $\mathbf{f}_{C}$ Embedding vs.~Unseen Categories} Fig.~\ref{fig:latent_} (a,b) shows t-SNEs of $\mathbf{f}_{S}$ and $\mathbf{f}_{C}$ on $13$ categories. We observe that $\f_C$ is more  discriminative, allowing the shape decoder to capture more intra-class deformations.
Further, we explore how well our \textbf{shape decoder} can represent the $3$D shape of {\it unseen classes}. 
Thus, we randomly select $20$ samples from each of $8$ unseen ShapeNet categories. 
%
With the sampled point-value pairs of each shape, we optimize its $\mathbf{f}_{C}$ and $\mathbf{f}_{S}$ via back-propagation through our trained shape decoder. As in Fig.~\ref{fig:latent_} (d,e), our reconstructions closely match the ground-truth. Quantitatively, we achieve a promising CD on unseen categories compare to that of unseen samples of seen categories:  $0.209$ vs.~$0.135$. Additionally, we ablate our decoder with or without category code: $0.209$ vs.~$0.267$, which demonstrates $\f_C$ enhances generalizability to unseen categories.  We further visualize the estimated $\mathbf{f}_{C}$ together with all training samples in Fig.~\ref{fig:latent_} (c). As we can see, $\mathbf{f}_{C}$ of unseen classes do not overlap with any training categories.

\begin{table}[t]
\centering
\vspace{0mm}
\caption{\small Effect of loss terms on pose estimation and reconstruction.}
\vspace{0mm}
\centering
\resizebox{1\linewidth}{!}{
\begin{tabular}{l | c| c |c | c }
\hline
& w/o $\mathcal{L}_{\text{sil}}$ & w/o $\mathcal{L}_{\text{fea-const}}$ & w/o $\mathcal{L}_{\text{reg}}$ & Full model \\
\hline 
\hline 
Azimuth angle error $\downarrow$& $17.89^{\circ}$ & $15.31^{\circ}$ & $13.32^{\circ}$ & $11.56^{\circ}$\\
Reconstruction (CD) $\downarrow$&  $0.145$ &  $0.133$ &   $0.137$  &   $0.113$ \\
\hline 
\end{tabular}
}
\label{tab:feature_loss}
\vspace{-2mm}
\end{table}

\Paragraph{Effect of Loss Terms}
Using car images of Pascal $3$D+, we compare our full model with its partial variants, in term of pose estimation and reconstruction (Tab.~\ref{tab:feature_loss}). 
As the silhouette provides strong constraints on global shape and pose, without silhouette loss, performance on both metrics are severely impaired. 
The regularization helps to disentangle shading from albedo, which leads to better surface normal, thus better shape and pose. 
The local feature consistency loss helps to fine-tune the model fitting, which improves the final pose and shape estimation. 
Thus all the loss terms in real data training contribute to the final performance.

\Paragraph{Effect of Training on Real Data} 
We evaluate $3$D reconstructions on images from Pix$3$D and Pascal $3$D${+}$ using models obtained at different training steps. The model fine-tuned on real images (\textbf{Pro.} (real)) has lower Chamfer distances compare to the model learned without real images (\textbf{Pro.}) for every single category (Tab.~\ref{tab:pascal_pix3d_rec}). 

\begin{figure}[t]
\centering
\includegraphics[trim=0 0 0 10, clip,width=\linewidth]{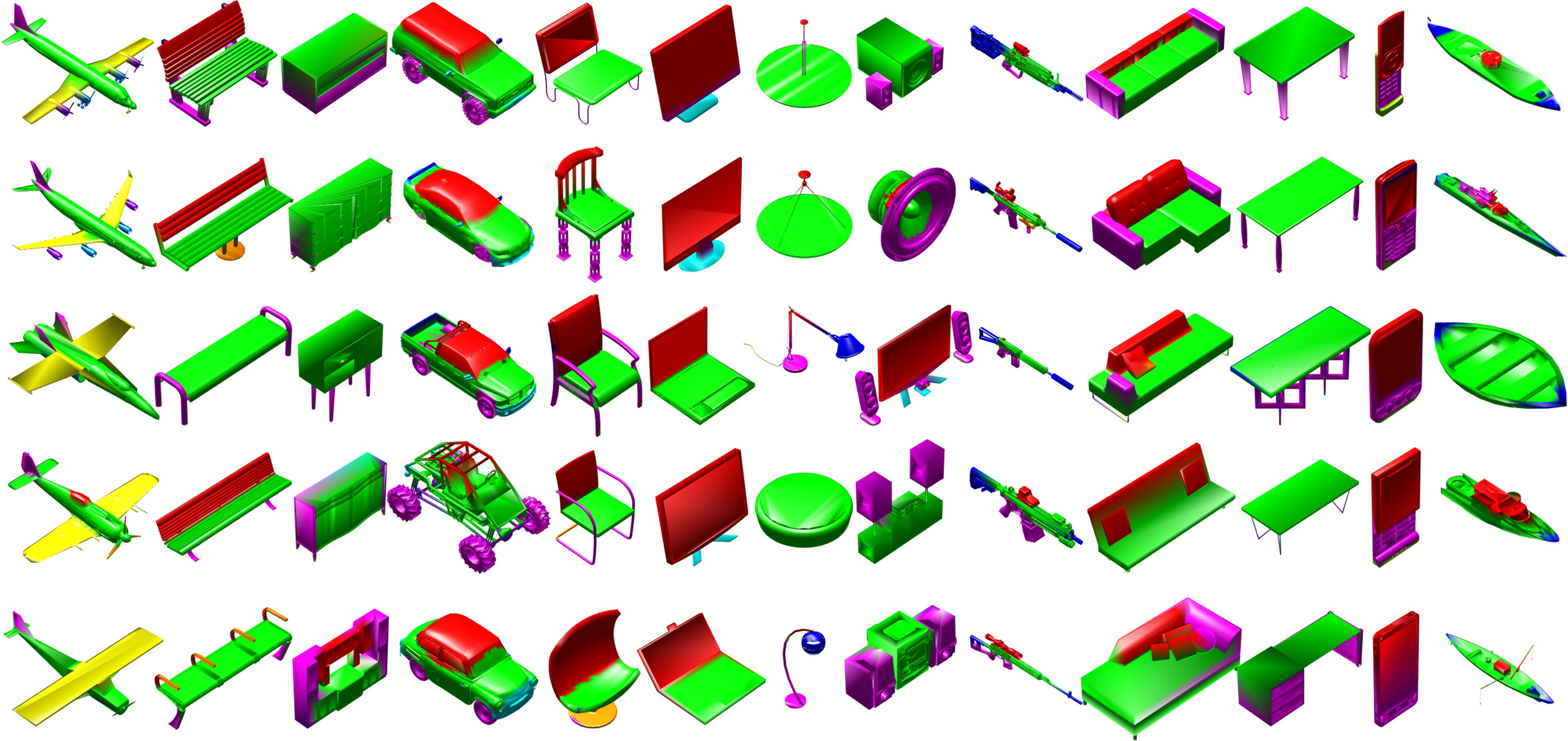}
\vspace{-3mm}
\caption{\small Unsupervised co-segmentation across $13$ categories.} 
\label{fig:co_seg}
\vspace{-2mm}
\end{figure}

\SubSection{Unsupervised Segmentation}
As modeling shape, albedo and co-segmentation are closely related tasks~\cite{taskonomy2018}, joint modeling allows exploiting their correlation. 
Following the same setting as~\cite{chen2019bae}, we evaluate CS models' co-segmentation and shape representation power on categories of airplane, chair and table. As in Tab.~\ref{tab:segmentation}, our model achieves a higher segmentation accuracy than BAE-NET~\cite{chen2019bae}. 
Further, we compare the ability of two models in representing $3$D shapes. 
By feeding a ground-truth voxel from the testing set to the voxel encoder $\Enc'$ and then shape decoder $\Dec_S$, we evaluate how well the shape-parameter-decoded shape matches the ground-truth CAD model. The higher IoU and lower CD show that we improve both segmentation and representation accuracy.
Further, Fig.~\ref{fig:co_seg} shows the co-segmentation across $13$ categories by our SU model. 
Meaningful segmentation appears both {\it within} a category and {\it across} categories.
For example, chair seats, plot in green, consistently correspond to sofa seats, table tops, and bodies of airplane, car and watercraft.  

\begin{table}[t!]
\centering
\newcommand{\tabincell}[2]{\begin{tabular}{@{}#1@{}}#2\end{tabular}}
\vspace{-2mm}
\caption{\small Segmentation/shape representation on ShapeNet part~\cite{yi2016scalable} in IoU$\uparrow$/CD$\downarrow$. The results are based on CS models without $\f_C$. 
}
\vspace{0.5mm}
\centering
\resizebox{1\linewidth}{!}{
\begin{tabular}{l| c| c| c| c }
\hline
 \tabincell{c}{Shape (\#parts)} &
 \tabincell{c}{airplane ($3$)} &
 \tabincell{c}{chair ($3$)} &
 \tabincell{c}{chair{+}table ($4$)} &
 \tabincell{c}{table ($2$)} \\
\hline 
\hline 
BAE-Net~\cite{chen2019bae}  &  $80.4/0.14$ & $86.6/0.18$ & $83.7/-$ & $87.0/0.16$  \\
Proposed &   $83.0/0.12$  & $87.4/0.15$ &    $84.1/0.14$   & $88.2/0.13$  \\
\hline 
\end{tabular}
}
\label{tab:segmentation}
\vspace{0mm}
\end{table}


\SubSection{Single-view $\mathbf{3}$D Reconstruction}


\begin{figure*}[t]
\centering     
\subfigure[]{\includegraphics[trim=0 28 0 2,clip, height=58mm]{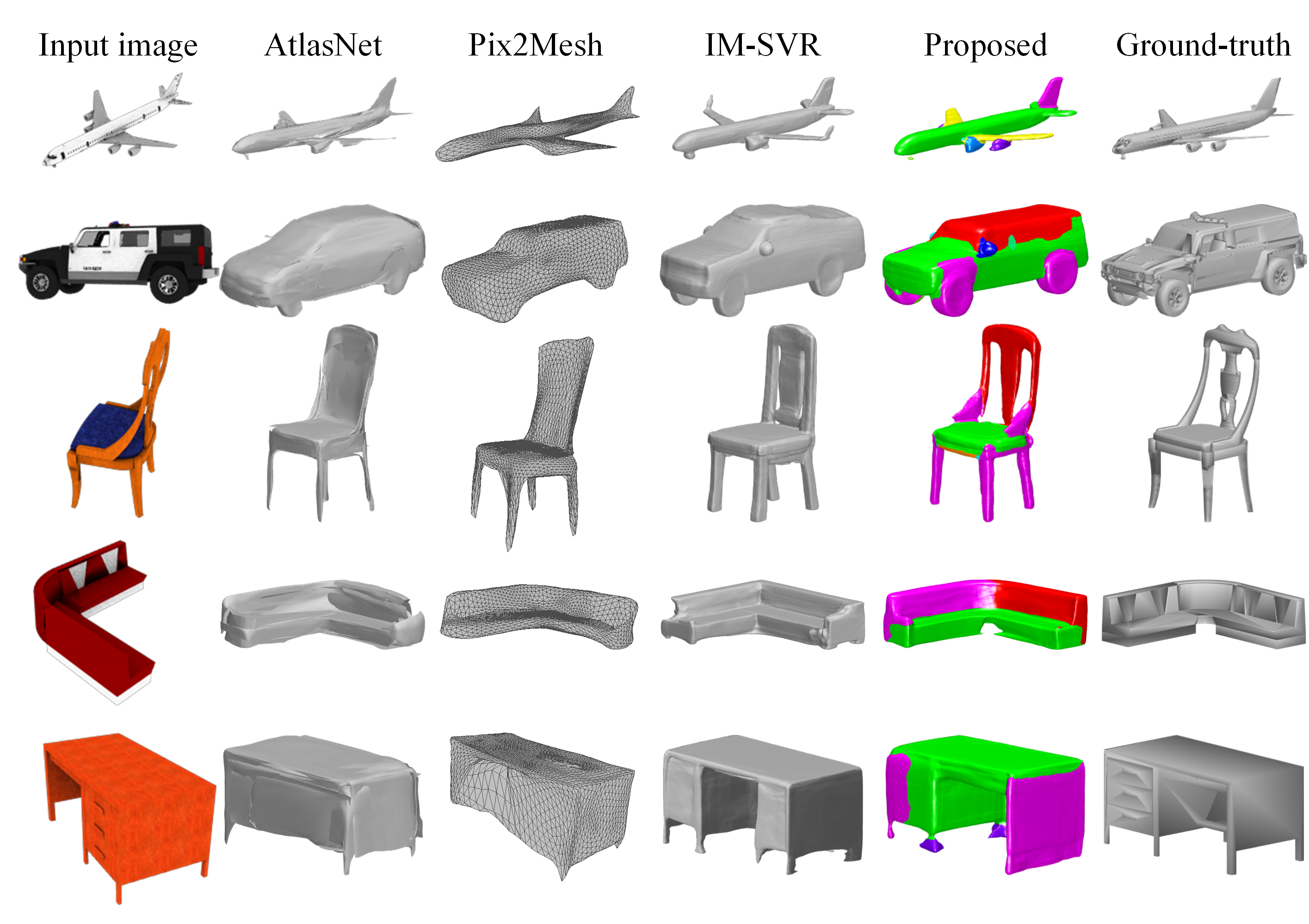}} 
\hspace{0.5mm}
\subfigure[]{\includegraphics[trim=0 15 0 2,clip,height=57mm]{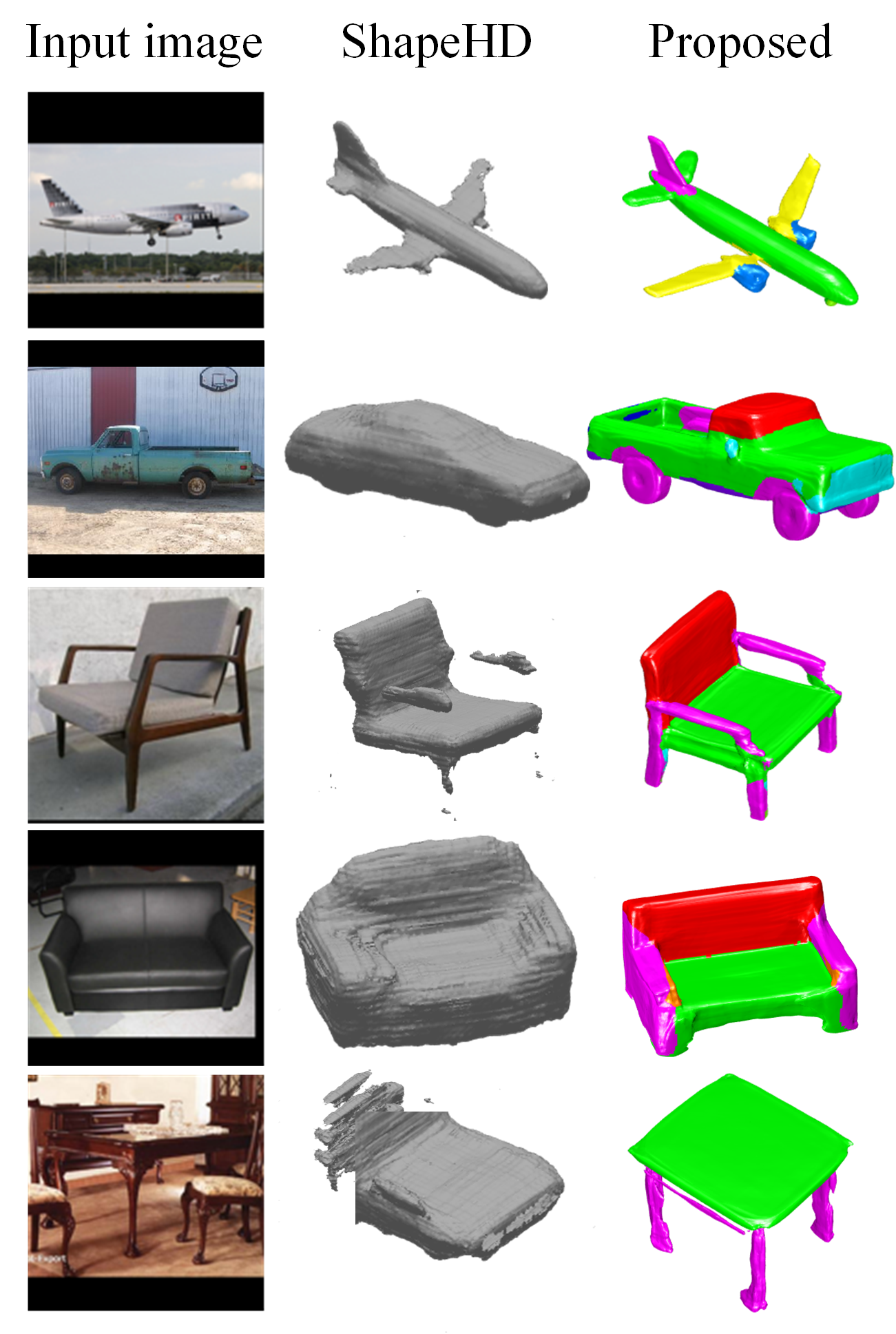}}
\hspace{0.5mm}
\subfigure[]{\includegraphics[trim=0 15 0 0,clip,height=57mm]{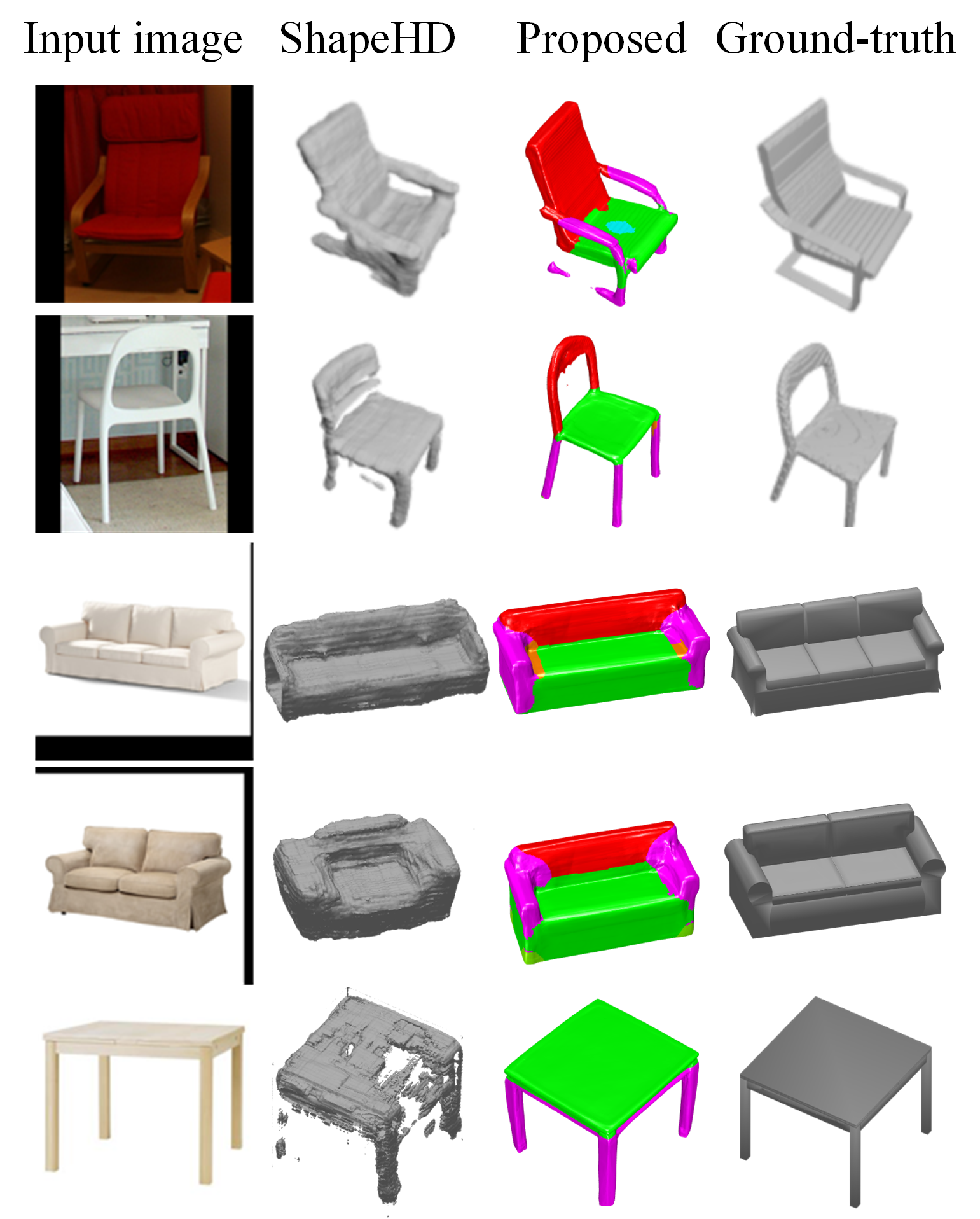}}
\vspace{-2mm}
\caption{\small Qualitative comparison for single-view $3$D reconstruction on (a) ShapeNet, (b) Pascal $3$D+, and (c) Pix3D datasets.
}
\label{fig:rec}
\vspace{-1mm}
\end{figure*}

\begin{table*}[t]
\centering 
\newcommand{\tabincell}[2]{\begin{tabular}{@{}#1@{}}#2\end{tabular}}
\caption{\small Quantitative comparison of $3$D reconstruction on synthetic images of ShapeNet. [Key: \firstkey{Best}, \secondkey{Second Best}] 
}
\resizebox{1\linewidth}{!}{
\begin{tabular}{l| c c c c c c c | c | c c c c c c| c }
\hline

\multirow{2}{*}{\textbf{Category}} &
\multicolumn{8}{c|}{Chamfer-$L_{1}$ Distance $\downarrow$} & 
\multicolumn{7}{c}{F-score (\%, $\tau{=}10^{-4}$) $\uparrow$} \\
\cline{2-8}
\cline{9-16}
& \tabincell{c}{$3$D-R$2$N$2$ \\ \cite{choy20163d}}
& \tabincell{c}{PSG \\ \cite{fan2017point}}
& \tabincell{c}{Pix$2$Mesh \\ \cite{wang2018pixel2mesh}}
& \tabincell{c}{AtlasNet \\ \cite{groueix2018atlasnet}}
& \tabincell{c}{IM-SVR \\ \cite{chen2018learning}}
& \tabincell{c}{ONet \\ \cite{mescheder2018occupancy}}
& \tabincell{c}{F2B \\ \cite{yao2020front2back}}
&\textbf{Pro.} 
& \tabincell{c}{$3$D-R$2$N$2$ \\ \cite{choy20163d}}
& \tabincell{c}{PSG \\ \cite{fan2017point}}
& \tabincell{c}{Pix$2$Mesh \\ \cite{wang2018pixel2mesh}}
& \tabincell{c}{AtlasNet \\ \cite{groueix2018atlasnet}}
& \tabincell{c}{IM-SVR \\ \cite{chen2018learning}}
& \tabincell{c}{F2B \\ \cite{yao2020front2back}}
& \textbf{Pro.}\\

\hline
\hline 
firearm       & $0.183$ & $0.134$ & $0.164$ & $\secondkey{0.115}$ & $0.126$ & $0.141$  & $0.127$ & $\firstkey{0.113}$   & $28.34$ & $69.96$ & $73.20$ & $75.98$ & $\firstkey{81.35}$ &  $76.90$ &  $\secondkey{79.56}$\\ 

car           & $0.213$ & $0.169$ & $0.180$ & $0.141$ & $\secondkey{0.123}$ & $0.159$ & $0.161$ & $\firstkey{0.115}$  & $37.80$ & $50.70$ & $67.86$ & $66.72$ & $\firstkey{75.89}$ &  $68.30$ &  $\secondkey{75.68}$\\ 

airplane      & $0.227$ & $0.137$ & $0.187$ & $\firstkey{0.104}$ & $0.137$ & $0.147$ & $0.127$ & $\secondkey{0.123}$ & $41.46$ & $68.20$ & $71.12$  & $70.22$ &  $\firstkey{79.15}$ &  $\secondkey{77.47}$ & $74.86$ \\ 

cellphone     & $0.195$ & $0.161$ & $0.149$ & $0.128$ & $\secondkey{0.131}$& $0.140$  & $0.135$ & $\firstkey{0.130}$  & $42.31$ & $55.95$ & $70.24$ & $71.97$ & $71.27$ &  $\firstkey{77.15}$ &  $\secondkey{73.91}$\\ 

bench         & $0.194$ & $0.181$ & $0.201$ & $\secondkey{0.138}$ & $0.173$& $0.155$ & $0.177$ & $\firstkey{0.137}$ & $34.09$  & $49.29$ & $57.57$  & $65.31$ & $65.60$ & $\firstkey{66.59}$ &  $\secondkey{66.15}$\\ 

watercraft    & $0.238$ & $0.188$ & $0.212$ & $\secondkey{0.151}$ & $0.157$& $0.218$ & $0.171$ & $\firstkey{0.143}$  & $37.10$ & $51.28$ & $55.12$ & $\firstkey{67.30}$ & $\secondkey{63.15}$ &  $63.04$ &  $60.90$\\ 

chair         & $0.270$ & $0.247$ & $0.265$ & $0.209$ & $0.199$ & $0.228$ & $\secondkey{0.184}$  &$\firstkey{0.160}$  & $40.22$ & $41.60$ & $54.38$ & $57.62$ & $62.41$ &  $\firstkey{64.72}$ &  $\secondkey{63.24}$\\ 

table         & $0.239$ & $0.222$ & $0.218$ & $0.190$ & $0.173$ & $0.189$ & $\firstkey{0.167}$ & $\secondkey{0.172}$  & $43.79$ & $53.44$ & $66.30$ & $69.49$ & $70.33$ &  $\firstkey{74.80}$ &  $\secondkey{71.27}$\\ 

cabinet       & $0.217$ & $0.215$ & $0.196$ & $0.175$ & $0.198$& $\firstkey{0.167}$ & $0.238$ & $\secondkey{0.174}$  & $49.88$ & $39.93$ & $60.39$  & $55.95$ & $\firstkey{68.42}$ &  $56.64$ &  $\secondkey{64.79}$\\ 

couch         & $0.229$ & $0.224$ & $0.212$ & $\firstkey{0.177}$  & $0.194$ & $0.194$ & $0.209$ &  $\secondkey{0.186}$ & $40.01$ & $36.59$ & $51.90$ & $52.61$ & $59.93$ & $\secondkey{61.59}$ &  $\firstkey{62.01}$\\ 

monitor       & $0.314$ & $0.284$ & $0.239$ & $\secondkey{0.198}$ & $0.225$ & $0.278$   & $\firstkey{0.185}$ & $0.208$  & $34.38$ & $40.53$ & $51.39$ & $56.55$ & $59.42$ &  $\secondkey{63.03}$ &  $\firstkey{71.45}$\\ 

speaker       & $0.318$ & $0.316$ & $0.285$ & $\secondkey{0.245}$ & $0.252$ & $0.300$  & $\firstkey{0.227}$ & $0.245$ & $45.30$ & $32.61$ & $48.84$ & $48.63$ & $56.87$ &   $\secondkey{59.10}$ & $\firstkey{63.19}$\\

lamp          & $0.778$ & $0.314$ & $0.308$ & $0.305$ & $0.362$ &  $0.479$  & $\firstkey{0.209}$ & $\secondkey{0.276}$  & $32.35$ & $41.40$ & $48.15$ & $57.42$ & $56.18$ &  $\firstkey{65.11}$ & $\secondkey{63.38}$\\

\hline 
\textbf{Mean} & $0.278$ & $0.188$ & $0.216$ & $\secondkey{0.175}$ & $0.187$ &  $0.215$ & $0.178$  &$\firstkey{0.168}$ & $39.01$ & $48.58$ & $59.72$ & $62.75$ & $66.92$ &  $\secondkey{67.26}$ &  $\firstkey{68.49}$\\
\hline 
\end{tabular}
}
\label{tab:syn_rec}
\vspace{-2mm}
\end{table*}

\vspace{1mm}
\Paragraph{Synthetic Images} 
We first evaluate $3$D reconstruction on synthetic images. 
We compare with SOTA baselines that leverage various $3$D representations: $3$D-R$2$N$2$~\cite{choy20163d} (voxel), Point Set Generation (PSG)~\cite{fan2017point} (point cloud), Pixel2Mesh~\cite{wang2018pixel2mesh}, AtlasNet~\cite{groueix2018atlasnet}, Front2Back~\cite{yao2020front2back} (mesh), and IM-SVR~\cite{chen2018learning}, ONet~\cite{mescheder2018occupancy} (implicit field). All baselines train a single model on $13$ categories, except IM-SVR which learns $13$ models. 
We report the results of our SU model, trained only on synthetic images,  without Step $3$. 


In general, our model is able to predict $3$D shapes that closely resemble the ground truth~(Fig.~\ref{fig:rec} (a)). 
Our approaches outperform baselines in most categories and achieves the best mean score, in both CD and F-score (Tab.~\ref{tab:syn_rec}).
While using the same shape representation as ours, IM-SVR~\cite{chen2018learning} only learns to reconstruct $3$D shapes by minimizing the latent representation difference with ground-truth latent codes. 
By modeling albedo, our model benefits from learning with both supervised and self-supervised (photometric, silhouette) losses. 
This results in better performance both quantitatively and qualitatively.

\Paragraph{Real Images} 
We evaluate $3$D reconstruction on two real image databases, Pascal~$3$D${+}$~\cite{xiang2014beyond} and Pix$3$D~\cite{sun2018pix3d} (overlapped categories only). 
We report two results of our method: a model trained with synthetic data only (\textbf{Pro.}) and a model fine-tuned on real images of Pascal~$3$D${+}$~\textbf{\emph{train}} subset \emph{without} access to ground truth $3$D shapes (\textbf{Pro.} (real)). 
Baselines include SOTA methods performed well on real images: $3$D-R$2$N$2$~\cite{choy20163d}, DRC~\cite{tulsiani2017multi}, ShapeHD~\cite{wu2018learning} and DAREC~\cite{pinheiro2019domain}. 
Among them, DRC and DAREC were trained on real images of Pascal~$3$D${+}$ as they adopt a differentiable geometric consistency or domain adaptation in training. 
3D-R2N2 and ShapeHD  cannot be fine-tuned on real images, without albedo modeling and rendering layer.
%

As in Fig.~\ref{fig:rec} (b), our model infers reasonable shapes even in challenging conditions. Quantitatively, Tab.~\ref{tab:pascal_pix3d_rec} shows that both proposed models outperforms other methods in Pascal~$3$D${+}$.
The clear performance gap between our two models shows the importance of training on real data.

As Pascal $3$D${+}$ only has $10$ CAD models per category as ground truth shapes, ground truth labels may be inaccurate. 
We therefore conduct experiments on Pix$3$D database with more precise $3$D labels. As in Tab.~\ref{tab:pascal_pix3d_rec}, our fine-tuned model has significantly lower CD and the best quality in Fig.~\ref{fig:rec} (c) comparing to baselines, which indicates our method can leverage real-world images without $3$D annotations via self-supervised learning. 

\begin{table}[t]
\newcommand{\tabincell}[2]{\begin{tabular}{@{}#1@{}}#2\end{tabular}}
\caption{\small Real $3$D reconstruction (CD $\downarrow$) on Pascal $3$D${+}$ and Pix$3$D.}
\vspace{0mm}
\centering
\resizebox{\linewidth}{!}{
\begin{tabular}[]{l l c c c c | c c}
\hline
 & 
& \tabincell{c}{$3$D-R$2$N$2$ \\ \cite{choy20163d}}
& \tabincell{c}{DRC \\ \cite{tulsiani2017multi}}
& \tabincell{c}{ShapeHD \\ \cite{wu2018learning}}
& \tabincell{c}{DAREC \\ \cite{pinheiro2019domain}}
  & \textbf{Pro.} & \tabincell{c}{\textbf{Pro.} (real)}\\ 
\hline
\hline
\parbox[t]{1mm}{\multirow{6}{*}{\rotatebox[origin=c]{90}{Pascal $3$D{+}}}}

& plane      & $0.305$ & $0.112$ & $\firstkey{0.094}$ & $0.108$ & $0.114$ & $\secondkey{0.102}$  \\
& car        & $0.305$ & $\firstkey{0.099}$ & $0.129$ & $\secondkey{0.100}$ & $0.128$ &$0.113$  \\
& chair      & $0.238$ & $0.158$ & $0.137$ & $\secondkey{0.135}$ & $0.138$   &$\firstkey{0.119}$ \\
& table      & $0.321$ & $0.162$ & $\secondkey{0.153}$ & -       & $0.167$  &$\firstkey{0.127}$  \\
& couch      & $0.347$ & $0.169$ & $0.176$ & -       & $\secondkey{0.157}$  &$\firstkey{0.138}$  \\

\cline{2-8}
& \textbf{Mean} & $0.303$ & $0.140$ & $\secondkey{0.138}$ & - & $0.141$ & $\firstkey{0.120}$ \\
\hline
\hline
\parbox[t]{1mm}{\multirow{4}{*}{\rotatebox[origin=c]{90}{Pix$3$D}}}
& chair      & $0.239$ & $0.160$ & $0.123$  & $0.112$  &$\secondkey{0.102}$ &$\firstkey{0.091}$  \\
& couch      & $0.307$ & $0.178$ & $\secondkey{0.137}$  & -  & $0.142$ & $\firstkey{0.114}$   \\
& table      & $0.289$ & $0.163$ & $\secondkey{0.133}$  & -  & $0.145$   &$\firstkey{0.127}$  \\
\cline{2-8}
& \textbf{Mean} & $0.278$ & $0.167$ & $\secondkey{0.131}$ & - & $0.137$ &$\firstkey{0.111}$  \\
\hline
\end{tabular}
}
\label{tab:pascal_pix3d_rec}
\vspace{-4mm}
\end{table}

\section{Conclusions}\label{sec:con}
To better leverage real-world images in $3$D modeling, we present a semi-supervised learning approach jointly learns the models and the fitting algorithm.
While there still be a need of CAD models, our framework, with carefully-designed representation, architectures and loss functions, are able to effectively exploit real images in the training without $3$D ground truth.
%
%
Essentially, our method is applicable 
to any object category if both i) in-the-wild $2$D images and ii) CAD models are available.
We are interested in applying our method to a wide variety of object categories.

{\small
\bibliographystyle{ieee_fullname}
\bibliography{egbib}
}

\end{document}